%% file: main.tex
\documentclass[10pt,twocolumn,letterpaper]{article}

\usepackage[pagenumbers]{cvpr} %

\input{preamble}

\definecolor{cvprblue}{rgb}{0.21,0.49,0.74}
\usepackage[pagebackref,breaklinks,colorlinks,allcolors=cvprblue]{hyperref}

\def\confName{CVPR}
\def\confYear{2027}

\title{LiveView: Real-Time Camera-Controlled Re-filming of Streaming Videos}

\makeatletter
\renewcommand{\and}{\end{tabular}\hskip 2.5em\begin{tabular}[t]{c}}

\makeatother
\author{
Youcan Xu$^{1}$ \quad Xinjie Huang$^{1}$ \quad Jiaxin Shi$^{2}$ \quad Hengye Lyu$^{3}$ \quad
Zhen Wang$^{4}$ \quad Jun Xiao$^{1}$ \quad Long Chen$^{4}$ \\[0.5em]
{\normalsize $^{1}$ZJU \qquad $^{2}$Xmax.AI \qquad $^{3}$HKUST(GZ) \qquad $^{4}$HKUST}
}

\begin{document}
\twocolumn[{%
\renewcommand\twocolumn[1][]{#1}%
\maketitle
\vspace{-1.2em}
\begin{center}
    {\large\url{https://xyc-fly.github.io/LiveView/}}
\end{center}
\vspace{0.2em}
\begin{center}
    \centering
    \includegraphics[width=0.95\textwidth]{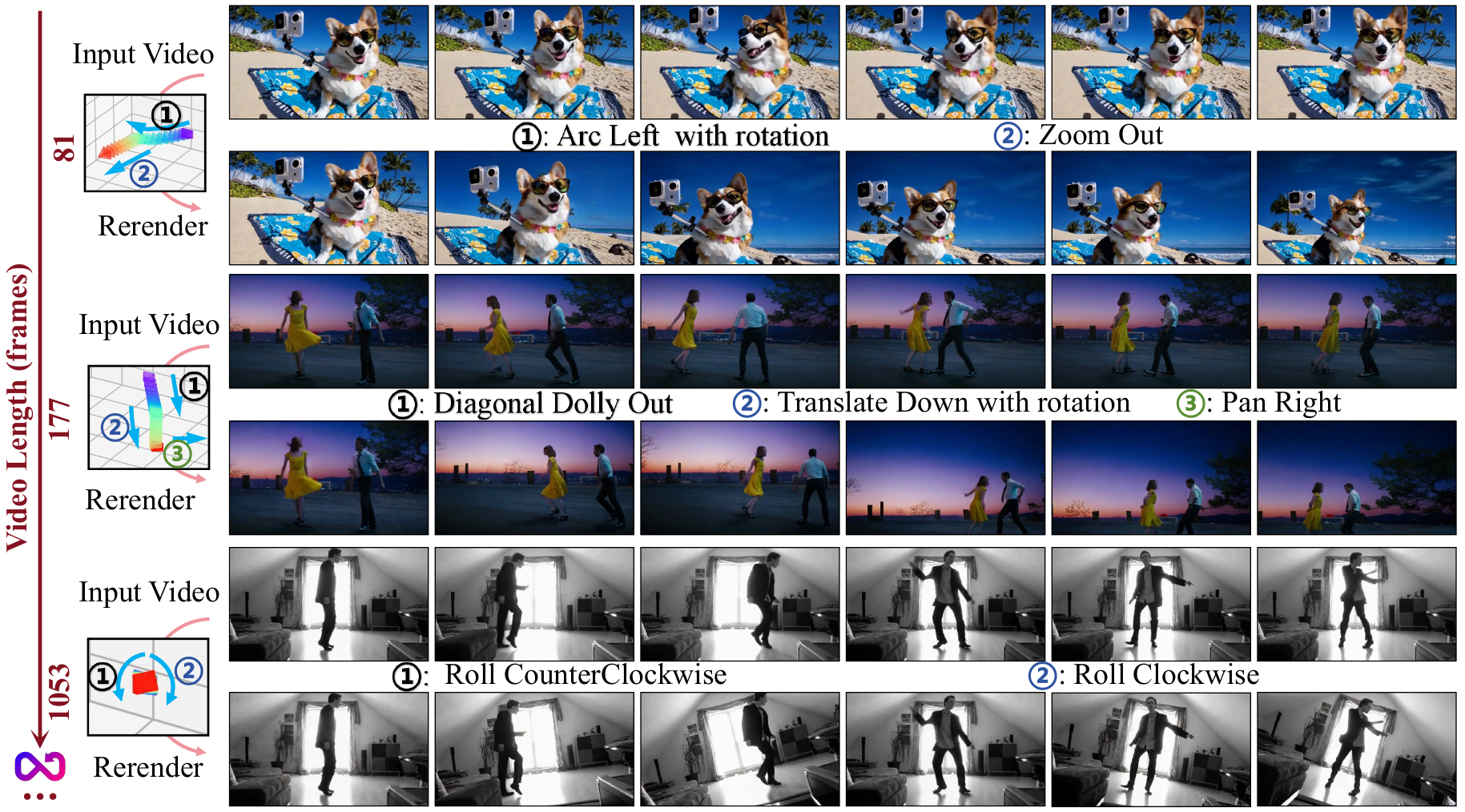}
    \captionof{figure}{\textbf{LiveView for real-time video re-filming.}
    Given a streaming input video and online user-specified camera
    trajectories (numbered arrows), LiveView enables real-time streaming
    re-filming of the dynamic scene from the requested viewpoints, preserving synchronization and content consistency.
    Although trained only on short
    clips, its length-agnostic streaming design imposes no fixed upper bound on
    rollout length---in principle supporting indefinitely long horizons; empirically, we show qualitative rollouts at 81, 177, and 1053 frames here.}
    \label{fig:teaser}
\end{center}
}]
\input{sec/0_abstract}
\input{sec/1_intro}
\input{sec/2_related}
\input{sec/3_method}

\input{sec/4_experiments}
\input{sec/5_conclusion}
{
    \small
    \bibliographystyle{ieeenat_fullname}
    \bibliography{main}
}

\end{document}

%% file: preamble.tex
\usepackage{colortbl}   %
\usepackage{makecell}   %
\usepackage{multirow}   %
\usepackage{algorithm}  %
\usepackage{algpseudocode}
\usepackage{tcolorbox}  %
\definecolor{Red}{RGB}{192, 0, 0}
\definecolor{Blue}{RGB}{12, 114, 186}

%% file: sec/0_abstract.tex
\begin{abstract}
Real-time camera-controlled video re-filming aims to synthesize target views from a streaming source video under online user-specified camera trajectories. This task poses two key challenges: maintaining synchronization and content consistency across views, and achieving low-latency rendering with long-horizon stability.  Existing approaches are predominantly designed for offline full-sequence processing and rely on bidirectional attention or costly depth-based geometric preprocessing, making them unsuitable for low-latency streaming scenarios. To address these limitations, we develop a teacher-to-student framework that distills a high-fidelity bidirectional re-filming teacher into a few-step causal autoregressive student. We introduce \emph{Cross-Frame Injection} to interleave each incoming source frame with its synchronized target latent, thereby establishing direct cross-view correspondence and improving content consistency. To ensure stable long-horizon generation, we further propose \emph{Shared Relative RoPE}, which bounds temporal indices within the pretrained range, and \emph{Cycle Streaming Tuning}, which exposes the student to extended cyclic trajectories during streaming training.
 These designs enable low-latency online video re-filming with precise camera control and sustained spatiotemporal consistency over long streaming rollouts.
\end{abstract}

%% file: sec/1_intro.tex
\section{Introduction}
 Interactive video world models~\citep{mao2025yume, parker2024genie, Hunyuan-gamecraft} have attracted increasing attention as a promising paradigm for understanding and interacting with the physical world. By modeling how dynamic environments evolve in response to external actions, such models provide a unified foundation for a wide range of applications, including embodied intelligence~\citep{kim2026cosmos}, autonomous driving~\citep{hu2023gaia},  and interactive content creation~\citep{Hunyuan-gamecraft, zhang2025matrix}. However, many existing interactive world models~\citep{mao2025yume} are
initialized from text prompts or static images, limiting their ability to ground predictions in an ongoing real-world scene. In contrast, interactive \emph{video-conditioned} world models~\citep{chen2025teleworld} continuously
integrate incoming video observations into an evolving world state and respond
to user actions in real time, enabling users or agents to interact with a
dynamically evolving environment. This capability is particularly
valuable for live broadcasting and filmmaking~\citep{luo2025camclonemaster, ICCV_recammaster}. For example, in
a live sports broadcast, a user steers a virtual camera around the players
as the footage streams in, and {expects to instantly view the action
from the requested viewpoint with minimal latency}. Realizing this vision demands real-time
responsiveness, precise camera control, and faithful preservation of the
incoming stream \emph{simultaneously}.

This motivates the task of real-time camera-controlled \emph{video re-filming} (\ie, re-rendering):
given a streaming source video that captures  the observed
world and online user-specified camera
trajectories, the model autoregressively synthesizes the corresponding
target views, rendering each new frame as the source stream advances.  This setting poses two fundamental
challenges: \textbf{1) Synchronization and content consistency across views}:
the model must faithfully preserve the content and dynamics
of the source stream while plausibly hallucinating regions that become newly
visible from the target viewpoints, all while precisely following the user's
camera controls.
\textbf{2) Low-latency rendering and long-horizon stability}: The target stream must be rendered with low latency and
maintain spatiotemporal consistency over long rollouts.

Despite the challenges outlined above, existing approaches rarely satisfy
these requirements simultaneously. Camera-controlled video re-rendering
methods~\citep{ICCV_recammaster, ICCV_TrajectoryCrafter, CVPR_redirector}
have achieved promising results in novel-view synthesis. However, they are
primarily designed for offline full-sequence processing: they either rely on
explicit geometric priors such as depth-based warping~\citep{ICCV_TrajectoryCrafter} or implicit camera
embeddings processed with bidirectional attention~\citep{ICCV_recammaster, CVPR_redirector}. Both paradigms require
the complete source context to be available before starting to generate,
making them poorly suited to low-latency interaction with incrementally
arriving video streams. More recently, Inspatio-World~\citep{team2026inspatio} introduces
a few-step causal autoregressive generator, together with depth-based
reference warping for autoregressive block-wise generation. Nevertheless, it still assumes a pre-observed
reference video and relies on offline feedforward reconstruction~\citep{depthA3}  to construct geometric priors. This pre-initialization requirement constitutes a fundamental limitation for real-time video-conditioned world modeling, where both scene observations and user controls must be handled online.

To address these limitations, we propose \emph{LiveView}, an
implicit framework for real-time camera-controlled video re-filming. Our
framework follows a teacher-to-student pipeline. We first train a high-fidelity
bidirectional camera-controlled video-to-video (re-filming) teacher, and then distill
it into a few-step causal autoregressive student for low-latency streaming
inference~\citep{dmd,yin2025Causvid,self_forcing}. Unlike Inspatio-World
~\citep{team2026inspatio}, which relies on time-consuming offline
pre-processing to align the source content, our framework operates directly
on the incoming source stream. Specifically, we introduce a
\emph{Cross-Frame Injection} mechanism that interleaves each incoming source
frame with its corresponding target latent as a synchronized contextual pair.
This design establishes direct cross-view correspondence at every time step,
improving temporal synchronization and content consistency with the source
stream.

However, we find that direct distillation~\citep{dmd} is insufficient to ensure
stable long-horizon streaming inference. We attribute this to two causes. \textbf{First}, the temporal RoPE indices grow with the global video time. Once the rollout
exceeds the horizon covered by pretraining, the model must extrapolate to
unseen positions, producing a positional train--test
mismatch~\citep{yesiltepe2025infinity}. \textbf{Second}, the student is typically trained on short
video clips, whereas streaming deployment requires it to autoregressively
generate substantially longer sequences. The resulting discrepancy in
sequence length can cause progressive degradation
over long rollouts.
To address these two issues, we introduce two complementary designs. 1) \textbf{Shared Relative RoPE (SR-RoPE)}. The key of SR-RoPE is to assign a shared temporal index to each
synchronized source-target frame pair. Rather than increasing with global
video time, the temporal indices are defined relative to the current cache window and
follow the same bounded schedule during training and inference. This keeps all
active temporal indices within the pre-trained range and thereby avoids the
extrapolation-induced mismatch. 2) \textbf{Cycle Streaming Tuning}. Starting from each short synchronized training
clip, we construct a long cyclic trajectory by extending the original video
and camera motion with their temporal reversals, causing the camera to return
to previously observed viewpoints. We expose the model to these long cyclic
sequences through efficient streaming training, thereby narrowing the gap
between short-clip training and long-horizon autoregressive inference.
Moreover, the cyclic construction enriches the camera trajectories.

Our contributions are threefold: 1) We propose \emph{LiveView}, a real-time
re-filming framework that operates directly on streaming input without
offline preprocessing. 2) We introduce \emph{Cross-Frame Injection} for synchronized
source conditioning, and \emph{Shared Relative RoPE} and \emph{Cycle
Streaming Tuning} to address positional extrapolation and error accumulation
in long-horizon re-filming. 3) LiveView achieves 10 FPS streaming with
2.76\,s startup latency, while maintaining competitive quality and camera
accuracy.

%% file: sec/2_related.tex
\section{Related Work}

\paragraph{Camera-Controlled Video Generation.}
With the goal of steering viewpoint changes through camera conditions,
camera-controlled video generation has become an important branch of controllable video synthesis~\citep{ma2025controllable_survey}. Early studies primarily explored this capability in text-to-video~\citep{he2024cameractrl,  kuang2024collaborative, AC3D, wang2025cinemaster, wang2026flowmotion} or image-to-video~\citep{viewcraft, Li_2025_ICCV, feng2024i2vcontrol,xu2024camco, Traj_attention, cheng2025adaptive, lyu2026dit} settings. More recently, growing attention has shifted toward camera-controlled video-to-video generation (re-rendering)~\citep{vanhoorick2024gcd,ICCV_recammaster, luo2025camclonemaster, li2026cameraanything}. This challenging task necessitates maintaining rigorous temporal synchronization and content consistency with the source video while precisely following the provided camera trajectory.
Current methods generally follow two technical routes. Warping-based methods \citep{Traj_attention, ICCV_TrajectoryCrafter, lu2025see4d, ReCapture, jeong2025reangle, lin2026vista4d} typically employ depth estimation to lift 2D frames into 3D point clouds, from which geometrically aligned proxy videos are rendered from target viewpoints. These proxies then serve as auxiliary conditions to guide the generative process.  Conversely, implicit synthesis methods \citep{ICCV_recammaster,CVPR_redirector, vanhoorick2024gcd} directly encode camera parameters into the model, often leveraging large-scale synthetic datasets for training. Despite their impressive visual quality, both paradigms are predominantly designed for offline, full-sequence modeling, rendering them computationally prohibitive for real-time interactive control.

\paragraph{Autoregressive Video Diffusion Models.}
Traditional video diffusion models~\citep{yang2024cogvideox, wang2025wan} typically rely on full-sequence, bidirectional modeling, which limits their scalability for long-horizon generation and real-time streaming. To overcome these constraints, a growing line of research~\citep{chen2024diffusion_forcing, Pyramidal,zhu2026causal,yuan2026helios} integrates autoregressive (AR) prediction with diffusion modeling. Early attempts like MAGI-1~\citep{teng2025magi} and CausVid~\citep{yin2025Causvid} transition from bidirectional to causal architectures, enabling chunk-wise rollout.
To mitigate the inherent challenges of temporal drift and error accumulation in AR sampling, recent works have introduced advanced training and inference strategies. For instance, Self-Forcing~\citep{self_forcing} addresses the train-test mismatch by conditioning the model on its own generated outputs. LongLive~\citep{yang2025longlive, chen2026longlive2} leverages attention-sink mechanisms and KV-cache management to maintain global consistency during extended rollouts. However, these methods are primarily optimized for prompt-based generation. While effective for visual quality, they often lack the flexibility required for fine-grained, user-driven control.

\paragraph{Interactive World Models.}
Interactive video world models~\citep{team2025hunyuanworld, wang2026matrix3, gao2026infinite}
simulate dynamic environments that evolve in response to user actions.
Recent work explores diverse control modalities, including camera poses~\citep{team2026advancing,sun2025worldplay} and
motion trajectories~\citep{zhao2025real_motion,shin2025motionstream}, audio~\citep{su2026omniforcing},
and text~\citep{yesiltepe2025infinity,li2025egoedit}.
However, many existing models are initialized from text prompts or static
images and generate subsequent dynamics without continuously incorporating
new observations from an ongoing scene. In this work, we bridge this gap by focusing on generating novel views of input videos through interactive camera control, enabling an immersive ``walk-through'' experience. Closest to our task, Inspatio-World~\citep{team2026inspatio}
supports real-time re-rendering but relies on offline geometric
preprocessing of a pre-observed reference video. LiveView instead
operates directly on incoming video observations and online camera controls,
enabling low-latency interaction without offline scene reconstruction.

%% file: sec/3_method.tex
\section{Approach}
\label{sec:approach}
\begin{figure*}[t]\centering
\includegraphics[width=0.95\textwidth]{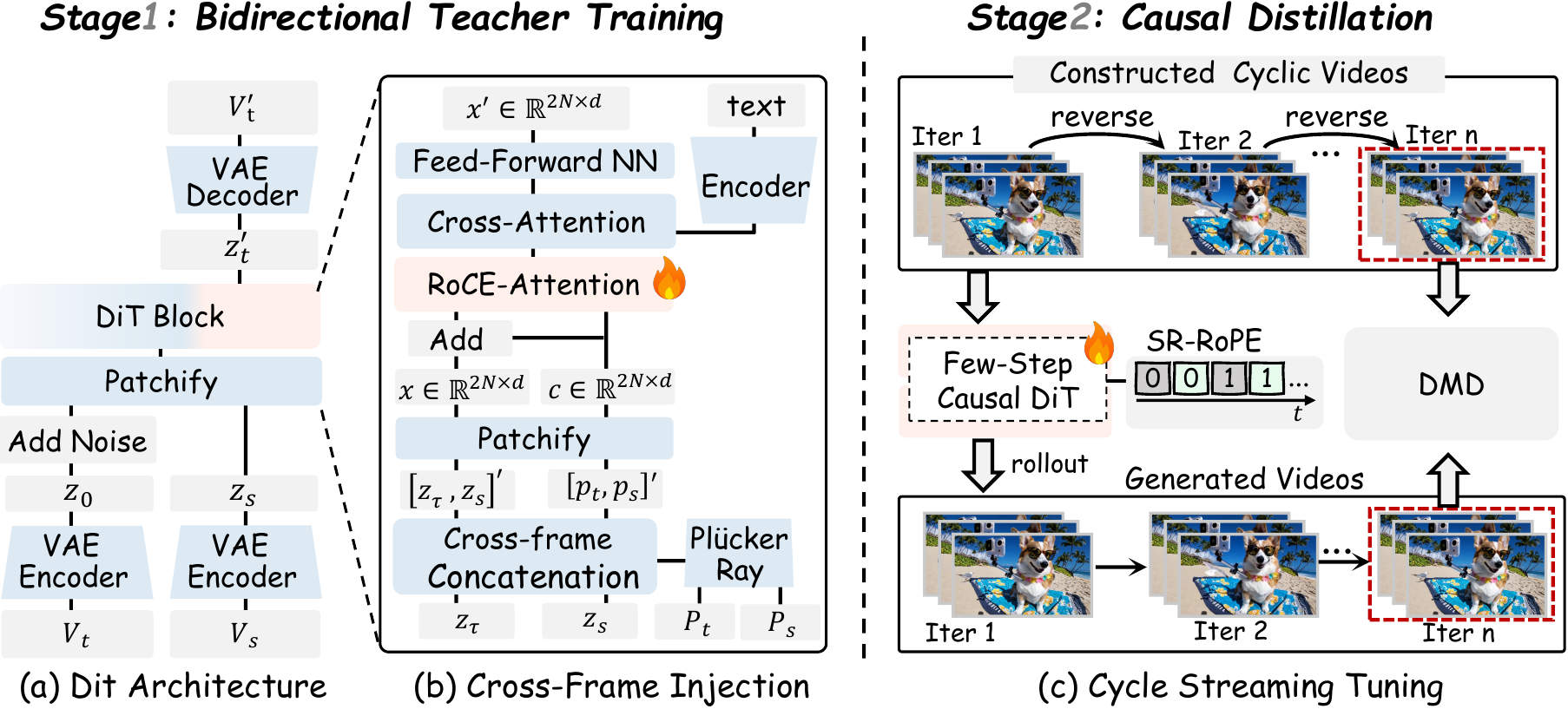}
\vspace{-.5em}
\caption{\textbf{Overview of the two-stage training pipeline of LiveView.} \textit{Stage 1:}  We train a bidirectional camera-controlled V2V teacher, where Cross-Frame Injection interleaves synchronized source--target latents and camera ray embeddings. \textit{Stage 2:} We distill the teacher into a few-step causal DiT via distribution matching distillation (DMD), enhanced with Cycle Streaming Tuning and Shared Relative RoPE (SR-RoPE) for stable long-horizon streaming generation.}
\label{fig_pipe}
\vspace{-1.5em}
\end{figure*}

As illustrated in Figure~\ref{fig_pipe}, LiveView follows a two-stage
training pipeline~\citep{yin2025Causvid,dmd}. We first train a bidirectional camera-controlled V2V
teacher (Sec.~\ref{sec:teacher}), and then distill it into a few-step
causal student for low-latency streaming inference (Sec.~\ref{sec:causal_student}). We further introduce
\emph{Shared Relative RoPE} and \emph{Cycle Streaming Tuning} to improve
long-horizon autoregressive stability.

\subsection{Preliminary}
\label{sec:preliminary}

\paragraph{3D Rotary Position Embedding (3D-RoPE).}
Following RoPE~\citep{rope}, the query ($q$) and key vectors ($k$) in self-attention layers are grouped into
complex-valued pairs, denoted as
$\bar{q},\bar{k}\in\mathbb{C}^{N\times (d_{\mathrm{head}}/2)}$, and rotated by
\begin{equation}
    \bar{q}'=\bar{q}\odot \mathbf{R},
    \qquad
    \bar{k}'=\bar{k}\odot \mathbf{R},
    \label{eq:rope_rotation}
\end{equation}
where $\mathbf{R}\in\mathbb{C}^{N\times (d_{\mathrm{head}}/2)}$ is the rotary matrix,
$\odot$ denotes element-wise
multiplication. The resulting attention logit between $n$-th query token 
and $m$-th key token is
\begin{equation}
    A(n,m)'
    =
\operatorname{Re}\!\left[\bar{q}'_n\bar{k}'_m{}^{*}\right]
    =
    \operatorname{Re}
    \left[
    \bar{q}_n
    \left(
    \bar{k}_m^{*}
    \odot
    e^{\mathrm{i}\theta_c(n-m)}
    \right)
    \right],
    \label{eq:rope_relative_attention}
\end{equation}
where $\operatorname{Re}[\cdot]$ extracts the real part and
$(\cdot)^{*}$ denotes complex conjugation, and
$\theta_c$ is the frequency related to the $c$-th rotary channel of matrix $\mathbf{R}$.
Current video diffusion transformers~\citep{yang2024cogvideox,wang2025wan} extend this construction to
the frame, height, and width axes by partitioning the rotary channels into
three groups. Accordingly, the 3D rotary matrix is constructed as
$\mathbf{R}=\mathbf{R}_f\oplus\mathbf{R}_h\oplus\mathbf{R}_w$, where
$\mathbf{R}_l\in\mathbb{C}^{N\times(d_{\mathrm{head}}/6)}$ denotes the
axis-specific rotation matrix for $l\in\{f,h,w\}$, and $\oplus$ denotes
channel-wise concatenation.
% \paragraph{Self-Forcing-Style Distillation.} 后面再管

\subsection{Camera-Controlled V2V teacher model}
\label{sec:teacher}
In this section, we fine-tune the pretrained Wan-I2V~\citep{wang2025wan,he2024cameractrl} bidirectional diffusion model into a high-fidelity
camera-controlled V2V teacher. Given a fixed-length source video
$V_s\in\mathbb{R}^{F\times C\times H\times W}$ and a user-specified target
camera trajectory $P_t$, the teacher generates a re-filmed video
$V_t\in\mathbb{R}^{F\times C\times H\times W}$ that follows $P_t$ while
preserving the content and dynamics of $V_s$. Our teacher contains two key components:
Cross-Frame Injection for synchronized source conditioning and
RoCE~\citep{CVPR_redirector} for camera conditioning.

 \paragraph{Cross-Frame Injection.}
Let $z_s=\mathcal{E}(V_s)$ and $z_0=\mathcal{E}(V_t)$ denote the source and
target latents encoded by the frozen video VAE, respectively. Following the
rectified-flow formulation~\citep{flux2024}, we sample $\epsilon\sim\mathcal{N}(0,I)$ and
$\tau\in[0,1]$ to obtain the noisy target latent $ z_\tau=(1-\tau)z_0+\tau\epsilon.$
Existing V2V models~\citep{ICCV_recammaster,luo2025camclonemaster} commonly concatenate the complete source and target
latents along the time. Under this organization, the distance
between corresponding source and target frames depends on the video length,
making their cross-view correspondence difficult to model.
We instead interleave the source and target latents at the frame level:
\begin{equation}
    \widetilde{z}_\tau
    =
   {[z_\tau,z_s]}^{'}
    =
    [z_\tau^1,z_s^1,z_\tau^2,z_s^2,\ldots,z_\tau^f,z_s^f],
    \label{eq:cross_frame_injection}
\end{equation}
where $f$ is the temporal length in the latent space. Each noisy target frame
is therefore placed adjacent to its synchronized clean source frame, forming
a local cross-view context pair. We further assign the same temporal and
spatial 3D-RoPE coordinates to the source and target tokens within each
pair. Consequently, their relative positional relationship remains unchanged
with the input length, allowing self-attention to directly associate each
target frame with the corresponding source observation. This organization
also provides a natural source--target unit for the subsequent causal streaming distillation. We validate this in our experiments in Table ~\ref{tab:ablation-177}.

\paragraph{Rotary Camera Encoding (RoCE).}
To provide precise camera control when generating dynamic scenes, we adopt RoCE~\citep{CVPR_redirector}\footnote{
We compare RoCE with other camera encodings in the 
supplementary.},
which injects camera information as relative phase shifts in the
self-attention layers. Specifically, let
$\mathbf{c}\in\mathbb{R}^{2N\times d}$ denote the Pl\"ucker-ray tokens
derived from the target and source camera trajectories, $P_t$ and $P_s$.
RoCE maps these tokens to camera-dependent phase shifts
$\phi_{\mathrm{qk}}=[\mathbf{0},
\operatorname{MLP}_{\mathrm{qk}}(\mathbf{c})]$ and constructs the camera
rotation matrix
$\tilde{\mathbf{R}}_{\mathrm{qk}}=e^{\mathrm{i}\phi_{\mathrm{qk}}}$.
The zero-padding leaves the temporal rotary channels unchanged, such that
only the spatial components encode camera information. The standard RoPE
transformation in Eq.~\ref{eq:rope_rotation} is then modified as:
\begin{equation}
    \bar{q}'=
    \bar{q}\odot\widetilde{\mathbf{R}}_{\mathrm{qk}}\odot\mathbf{R},
    \qquad
    \bar{k}'=
    \bar{k}\odot\widetilde{\mathbf{R}}_{\mathrm{qk}}\odot\mathbf{R}.
    \label{eq:roce_rotation}
\end{equation}
Consequently, the attention interaction  between tokens $n$ and $m$ becomes

\begin{equation}
\begin{gathered}
A'(n,m)
=\operatorname{Re}\!\Big[\bar{q}_n\Big(\bar{k}_m^{*}\odot
    e^{\mathrm{i}\left(\theta_c(n-m)+\Delta\phi_{\mathrm{qk}}\right)}\Big)\Big],\\
\Delta\phi_{\mathrm{qk}}=\phi_{\mathrm{qk}}(n,\mathbf{c})-\phi_{\mathrm{qk}}(m,\mathbf{c}),
\end{gathered}
\end{equation}
thereby jointly encoding relative positions and
camera poses.  We refer the reader
to \citep{CVPR_redirector} for the full derivation.

\subsection{Autoregressive DiT for Streaming Re-filming}
\label{sec:causal_student}
\begin{figure}[t]
    \centering
    \includegraphics[width=0.95\linewidth]{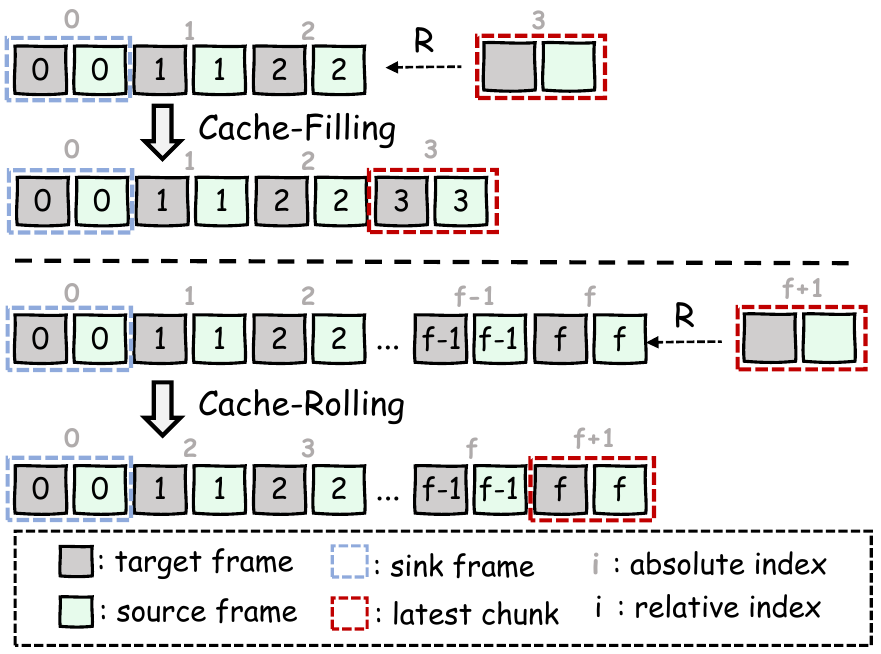}
    \vspace{-.5em}
    \caption{\textbf{Shared Relative RoPE.} Each source--target pair shares the
    same temporal index. Here, $f=f_{\max}$ denotes the maximum temporal index,
    and $\mathbf{R}$ denotes the rotary operation in Eq.~\ref{eq:rope_rotation}.}
    \label{fig:shared_rope}
\end{figure} 

Although the bidirectional teacher achieves high-fidelity re-filming, its
full-sequence attention and multi-step denoising preclude low-latency
streaming. We therefore distill it into a few-step causal student that renders
the target stream chunk by chunk. Following the Cross-Frame Injection design
in Eq.~\ref{eq:cross_frame_injection}, we define the $i$-th synchronized
chunk as $Z^i=[z_\tau^i,z_s^i]$, which pairs the noisy target latent to be
rendered with the clean source latent arriving online.
For each chunk, the student denoises only the target component while
conditioning on the source component and a bounded KV cache of previously
generated chunks. We optimize the student using Self-Forcing-style
distribution matching distillation~\citep{self_forcing,dmd}.
While the distilled student can perform autoregressive rollout, extending it
beyond the training horizon raises two challenges: (i) prediction errors
accumulate in the self-generated context, and (ii) temporal RoPE indices
drift into positional ranges unseen during training. We address them with
\emph{Cycle Streaming Tuning} and \emph{Shared Relative RoPE},
respectively.

\paragraph{Cycle Streaming Tuning.}
Causal distillation on short clips does not expose the student to the
progressively degraded self-generated contexts encountered during long
inference. Naively distilling an entire long rollout is also impractical:
the teacher is reliable only within its short training horizon, while
backpropagating through the complete sequence incurs prohibitive memory
cost~\citep{yang2025longlive}. We address both limitations with
\emph{Cycle Streaming Tuning} (CST), which constructs long synchronized
training streams from short V2V samples and applies teacher supervision only to a
local segment at the end of each rollout. As shown in Figure~\ref{fig_pipe} (c), given a training sample
$\mathcal{V}=(V_s,V_t,P_s,P_t)$ of $F$ frames, we synthesize an extended
stream by alternating between the original sequence and its temporally
reversed counterpart:
\begin{equation}
    \begin{gathered}
    \mathcal{V}^{\mathrm{long}}_n
    =
    \mathop{\bigoplus}_{j=0}^{n}\mathcal{V}^{(j)},
    \qquad
    n\sim\mathcal{U}\{0,\ldots,n_{\max}\},
    \\
    \mathcal{V}^{(j)}
    =
    \begin{cases}
        \mathcal{V}, & j \text{ is even},\\
        \mathcal{V}^{\mathrm{rev}}, & j \text{ is odd},
    \end{cases}
\end{gathered}
    \label{eq:cyclic_sequence}
\end{equation}
where $\bigoplus$ denotes temporal concatenation with duplicated turning
frames removed; the reversed sequence follows the temporal order
$\{F-1,F-2,\ldots,1\}$. The source video, target video, and their camera trajectories
are reversed jointly to preserve frame-level synchronization. Thus, $n$
controls the number of direction reversals: $n=0$ retains the original
sample, while larger $n$ produces progressively longer forward--backward
streams.

The student autoregressively processes the entire extended stream to build a
long self-generated history in its KV cache. We divide the stream into
fixed-length segments and stop gradients through all preceding segments. Distillation is applied
exclusively to the final segment, whose length remains same as the teacher's
training horizon. CST therefore exposes the student to realistic
long-horizon context degradation while retaining short-horizon teacher
supervision and bounded training memory. Beyond this efficient train-long--test-long optimization, the cyclic construction also increases
camera-trajectory diversity without requiring additional long-video data.

\paragraph{Shared Relative RoPE.}
A naive causal implementation assigns temporal indices according to the
global frame number. Once a rollout exceeds the pretrained range
$[0,f_{\max}]$, the model must evaluate RoPE phases at unseen indices,
resulting in quality degradation and eventual
overflow~\citep{yesiltepe2025infinity}. Training-free re-anchoring~\citep{yesiltepe2025infinity} at
inference can alleviate overflow, but leaves a positional train--test
mismatch as the model is still optimized using absolute indices.

We propose \emph{Shared Relative RoPE}, which applies the same
bounded, cache-relative schedule during both training and inference while
respecting the source--target pair structure. As shown in
Figure~\ref{fig:shared_rope}, temporal indices increase normally during the
\emph{Cache-Filling Phase}. Once the cache is full, the schedule enters the
\emph{Cache-Rolling Phase}: the oldest non-sink chunk is evicted, the newest
chunk is anchored at $f_{\max}$, and the remaining cached chunks are shifted
backward while the sink chunk remains fixed at the initial temporal index.
The temporal RoPE coordinates are therefore determined by each chunk's
relative position within the current cache window, rather than its global
frame index. Consequently, regardless of how long the global rollout becomes,
all active RoPE indices remain within $[0,f_{\max}]$, while the pairwise
temporal offsets within the rolling window are preserved.
Only the temporal component of 3D-RoPE is
rescheduled; the spatial coordinates and RoCE camera phases remain unchanged. 
In practice, the KV cache stores RoCE-conditioned but
RoPE-free keys, \eg,
$\bar{k}\odot\widetilde{\mathbf{R}}_{\mathrm{qk}}$ in
Eq.~\ref{eq:roce_rotation}. The cache-relative RoPE rotations are applied
only immediately before attention computation. This design preserves the camera
conditioning of previous frames directly in the KV cache, avoiding an
auxiliary cache for historical camera conditions. The complete procedure for our distillation incorporating our core designs
is detailed in Algorithm~1 of the supplementary material.

%% file: sec/4_experiments.tex
\section{Experiments}
\label{experiments}
\input{tables/Quantitative_all}
\begin{figure*}[t] 
    \centering
    \includegraphics[width=0.95\textwidth]{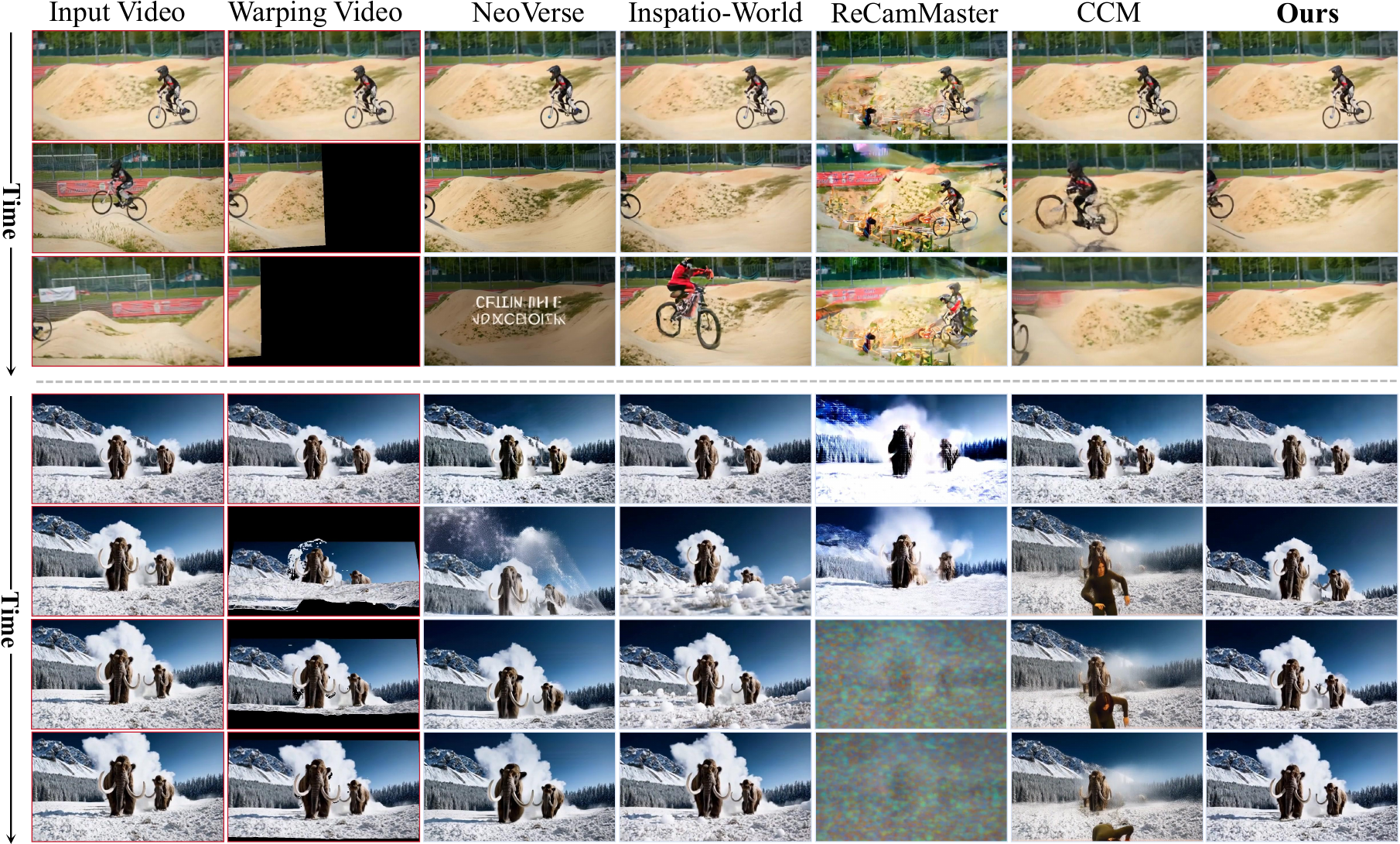}
    \vspace{-.5em}
    \caption{\textbf{Qualitative comparison with state-of-the-art methods.}
    The top is the short clip and the bottom is long clip.}
    \label{fig:qualitative_comparison}
\end{figure*}

\input{tables/ablation_main_177}
\begin{figure*}[t] 
    \centering
    \includegraphics[width=0.95\textwidth]{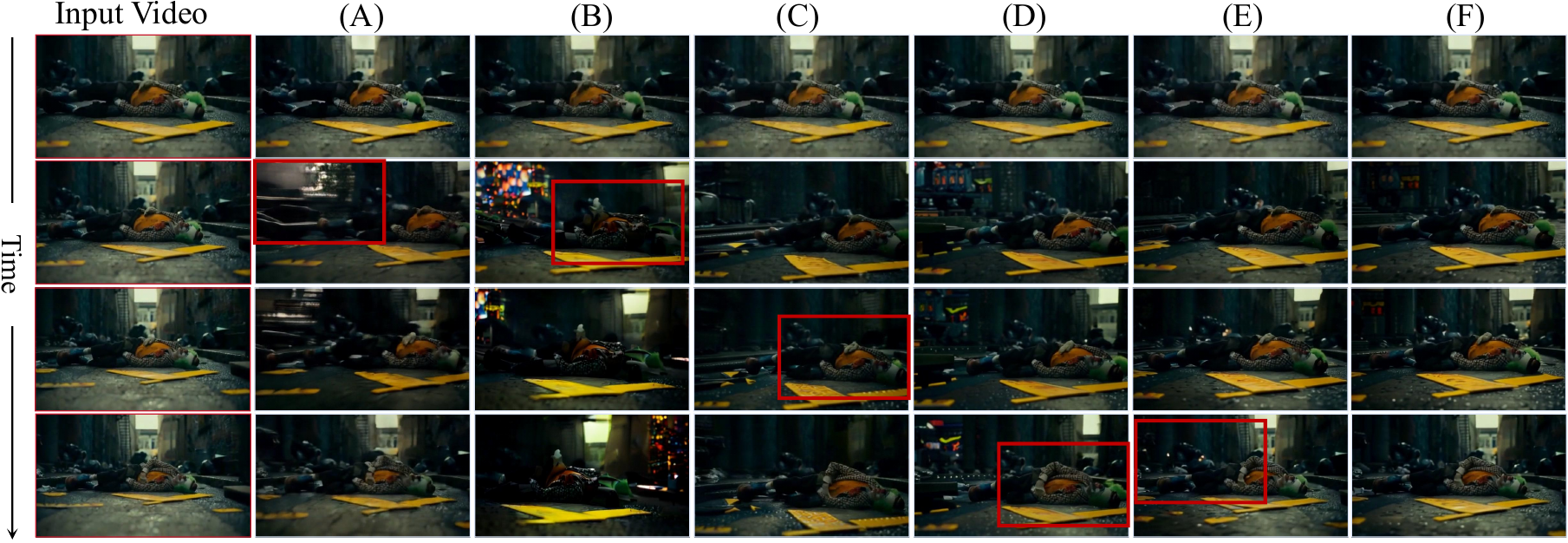}
    \vspace{-1.em}
    \caption{\textbf{Qualitative ablation study on 177-frame videos.}  \textcolor{red}{Red boxes} highlight failure cases.  Rows correspond to the settings in Table~\ref{tab:ablation-177}.  }
\label{fig:qual_ablation}
\end{figure*}

\paragraph{Implementation details.}
We first fine-tune the bidirectional teacher model on the MultiCamVideo dataset ~\citep{ICCV_recammaster} for 25K steps at a resolution of $81\times480\times832$ with a learning rate of 0.0001 and a batch size of 8. During teacher fine-tuning, we update only the self-attention layers and
keep all remaining modules frozen, as in prior camera-controlled V2V
models~\citep{ICCV_recammaster, CVPR_redirector}.  The causal student uses three denoising
steps (times steps $t_0 = 1000, t_1 =755, t_2 =522, t_3 = 0$), and all quantitative experiments use a KV cache size of 24 ( 12 latent pairs). We set chunk size to 3 pairs and $n_{max} = 2$. All the models are trained on four H20 GPUs. At inference, source camera poses are estimated online by the ViPE frontend~\citep{huang2025vipe} at 10 FPS on a separate GPU.

\paragraph{Baselines.}

We compare our method with state-of-the-art video re-rendering generation approaches. ReCamMaster~\citep{ICCV_recammaster}
and ReDirector~\citep{CVPR_redirector} use implicit camera conditioning,
while CamCloneMaster (CCM)~\citep{luo2025camclonemaster} transfers camera
motion from a reference video. TrajectoryCrafter~\citep{ICCV_TrajectoryCrafter},
NeoVerse~\citep{Neoverse_26}, and Inspatio-World~\citep{team2026inspatio}
use explicit geometric warping. Inspatio-World also employs a few-step
causal generator, making it a particularly relevant comparison for
streaming efficiency. Note its full system is not
streaming: it first reconstructs the entire pre-observed source video with
DA3~\citep{depthA3} before generation begins. 

\paragraph{Evaluation protocol.}
We construct two test sets with diverse real and synthetic content. The
short set contains 50 videos (33--81 frames) from DAVIS~\citep{davis} and
prior works~\citep{ICCV_recammaster, brooks-sora}; the long set contains 18
clips whose length is 177 frames.  We apply 8 distinct camera trajectories to each source
video, yielding 400 short and 144 long video--camera pairs, covering varied
content and motions while explicitly probing autoregressive stability over
extended rollouts. Note that these are natural continuous videos rather
than temporally reversed or looped sequences. For metrics, we report view synchronization with
CLIP-V~\citep{ICCV_recammaster}, video fidelity with the official VBench
suite~\citep{huang2024vbench}, and camera control with translation and
rotation error (Trans/Rot Err). We further report time-to-first-output
(latency) and FPS\footnote{The FPS reported by
Inspatio-World~\citep{team2026inspatio} excludes its feedforward
reconstruction, timing definitions for all methods
are given in the supplementary.\label{fn:inspatio_fps}} on H20 GPUs to assess real-time interactivity;
per-metric details and the user study are in the supplementary.

\subsection{Results}

\paragraph{Quantitative comparisons.}
Table~\ref{tab:main_quanti} shows the results on both test sets.
In terms of real-time interactivity, LiveView produces its first
output in 1.88\,s, comprising 0.9\,s of  pose initialization and 0.98\,s of
first-chunk generation, and subsequently sustains streaming rendering at
10 FPS. This reduces time-to-first-output by factors of approximately
$12$--$30$ relative to Inspatio-World, whose causal generator still requires offline
reconstruction of the full source video (23.6/57.2\,s).
Compared with offline methods, which require full-sequence processing
before producing output (62--4,140\,s), the latency reduction reaches
three orders of magnitude. Our throughput is also approximately
$4\times$ that of Inspatio-World (2.50--2.75 FPS).
In terms of quality, our model achieves the best subject consistency,
motion smoothness, and translation accuracy on short clips and maintains
competitive performance on 177-frame sequences. By comparison, some
implicit baselines degrade substantially over longer rollouts; for
example, ReCamMaster's CLIP-V score drops from 84.71 to 65.64.
Overall, LiveView
combines low startup latency and high streaming throughput with competitive
source fidelity and long-horizon consistency.

\paragraph{Qualitative comparisons.}
Figure~\ref{fig:qualitative_comparison} compares video re-filming under
diverse camera motions and sequence lengths. Warping-based methods~\citep{Neoverse_26, team2026inspatio}  exhibit
ghosting and structural distortions in regions where the reprojected source
becomes sparse or misaligned ( warping video in Figure~\ref{fig:qualitative_comparison}). Implicit methods~\citep{ICCV_recammaster, luo2025camclonemaster} remain plausible
on short clips but tend to lose view synchronization and produce blurry or
drifting content as the sequence extends beyond the training horizon. In
contrast, LiveView preserves sharper scene details, maintains temporal
synchronization with the source video, and follows the specified camera
trajectories in both short and long sequences.
Minute-scale results are shown in the supplementary material.

\subsection{Ablation Studies}

We ablate the core designs on 177-frame sequences in
Table~\ref{tab:ablation-177}, with qualitative visualizations shown in
Figure~\ref{fig:qual_ablation}. Additional ablation results are provided in the
supplementary.

\paragraph{Condition injection (A--B vs.\ F).}
Replacing Cross-Frame Injection with temporal concatenation (A)~\citep{ICCV_recammaster} places the
entire source sequence in a fixed prefix, weakening local cross-view
correspondence as the sequence grows and reducing synchronization and
consistency. Channel-wise concatenation (B)~\citep{shin2025motionstream} further lacks explicit temporal
alignment between corresponding source and target frames, resulting in poorer
view synchronization and camera accuracy. By contrast, Cross-Frame Injection
(F) maintains a synchronized source anchor for every target frame and achieves
the strongest overall performance.

\paragraph{Long-horizon designs (C--E vs.\ F).}
The directly distilled student (C) degrades over long rollouts as temporal
indices move beyond the pretrained range and errors accumulate in the
self-generated history. This degradation manifests as reduced camera accuracy
and visible temporal inconsistencies in Figure~\ref{fig:qual_ablation}. Adding
SR-RoPE alone (D) improves consistency and camera control by keeping all active
temporal indices within the pretrained range. Adding Cycle Streaming Tuning
alone (E) mainly improves image quality and camera accuracy by exposing the
student to extended self-generated contexts. Combining
SR-RoPE and CST (F) yields the best overall results, confirming that the two designs jointly improve
long-horizon autoregressive stability.
 

%% file: tables/Quantitative_all.tex
\begin{table*}[t!]
    \centering
    \caption{\textbf{Quantitative comparison with SOTA methods on short and long video clips.} 
    Our method maintains competitive visual quality and maintains accurate camera control while significantly reducing inference time.  
    Best results are in \textbf{bold} and second best are \underline{underlined}.}
    \vspace{-0.5em}
    \label{tab:main_quanti}
    \setlength\tabcolsep{2pt} 
    \resizebox{0.95\linewidth}{!}{%
    \begin{tabular}{@{}l*{13}{c}@{}}
       \toprule
        \multirow{2}{*}[-1.33ex]{Method}
        & \multicolumn{3}{c}{Latency (s) $\downarrow$}
        & \multicolumn{1}{c}{FPS}
        & \multicolumn{1}{c}{CLIP-V}
        & \multicolumn{6}{c}{VBench~\citep{huang2024vbench} $\uparrow$}
        & \multicolumn{2}{c}{Camera Acc. $\downarrow$} \\
        
        \arrayrulecolor{gray}
        \cmidrule(lr){2-4}
        \cmidrule(lr){7-12}
        \cmidrule(lr){13-14}
        
        & \makecell{Init.}  
        & \makecell{Gen.} 
        & \makecell{Total}
        & \makecell{$\uparrow$}
        & \makecell{$\uparrow$}
        & \makecell{Sub. \\ Cons}  
        & \makecell{Bg. \\ Cons.} 
        & \makecell{Aes. \\ Qual.}  
        & \makecell{Img. \\ Qual.} 
        & \makecell{Tem. \\ Flick.} 
        & \makecell{Motion \\ Smooth}  
        & \makecell{Trans \\ Err} 
        & \makecell{Rot \\ Err} \\
        \midrule
        
        \multicolumn{14}{c}{\textbf{Short Video Clips }} \\
        \midrule
        
        % GCD (2024) 
        % & -- & -- & -- & --
        % & 73.66
        % & 79.46 & 88.27 & 45.32 & 40.18 & 96.97 & 97.72 
        % & 0.09 & 12.6 \\

        ReCamMaster (2025) 
        & 0 & 426 & 426 & 0.19
        & 84.71
        & 89.49 & 93.08 & 52.58 & 53.40 & \underline{97.10} & 98.86 
        & 0.035 & 2.615 \\

        CamCloneMaster (2025) 
        & 0 & 1010 & 1010 & 0.08
        & 90.67
        & 90.37 & 93.07 & 54.08 & 56.13 & 96.63 & 99.01 
        & 0.0364 & 3.753 \\
        
        TrajectoryCrafter (2025) 
        & 74 & 495 & 569 & 0.14
        & 90.51
        & 91.58 & 94.04 & \underline{57.64} & 58.35 & \textbf{97.42} & 99.06 
        & 0.047 & 3.506 \\
        
        ReDirector (2026) 
        & 30 & 472 & 502 & 0.16
        & \underline{92.74}
        & \underline{93.30} & \underline{94.68} & 56.60 & 59.84 & 96.85 & \underline{99.13} 
        & \underline{0.0203} & \textbf{1.176} \\

        NeoVerse (2026) 
        & 12 & 50 & 62 & 1.31
        & 91.75
        & 93.16 & \textbf{94.69} & \textbf{57.79} & \textbf{63.68} & 96.35 & 99.09 
        & 0.0217 & 1.743 \\

        \arrayrulecolor{gray}\midrule
        
        {Inspatio-World (2026)} 
        & 22.8 & 0.8 & \underline{23.6} & \underline{2.75}
        & \textbf{93.01}
        & 92.86 & 94.40 & 55.27 & 59.29 & 96.63 & 98.95 
        & 0.024 & 1.665 \\
        
        \rowcolor{gray!10}
        \textbf{Ours} 
        & 0.9 & 0.98 & \textbf{1.88} & \textbf{10}
        & \underline{92.74}
        & \textbf{93.70} & 93.95 & 55.48 & \underline{60.43} & 96.81 & \textbf{99.26} 
        & \textbf{0.0178} & \underline{1.530} \\
        
        \midrule
        
        \multicolumn{14}{c}{\textbf{Long Video Clips (177 frames)}} \\
        \midrule
        
        % GCD (2024) 
        % & -- & -- & -- & --
        % & 73.36
        % & 79.09 & 88.08 & 45.16 & 40.26 & 96.93 & 97.70 
        % & 0.142 & 11.540 \\

        ReCamMaster (2025) 
        & 0 & 1771 & 1771 & 0.10
        & 65.64
        & 70.88 & 88.73 & 38.55 & 31.63 & \underline{98.40} & 98.88 
        & 0.148 & 12.181 \\

        CamCloneMaster (2025) 
        & 0 & 4140 & 4140 & 0.04
        & 78.18
        & 80.41 & 89.27 & 48.34 & 49.27 & \textbf{98.61} & 98.98 
        & 0.136 & 4.826 \\
        
        TrajectoryCrafter (2025) 
        & 94 & 1736 & 1830 & 0.096
        & 86.22
        & 91.63 & 94.41 & 55.63 & 58.28 & 97.14 & 98.74 
        & 0.133 & 4.692 \\
        
        ReDirector (2026) 
        & 45 & 2105 & 2150 & 0.082
        & 93.86
        & \underline{93.82} & \textbf{95.24} & \underline{60.16} & 60.70 & 97.53 & 99.18 
        & \textbf{0.034} & 2.660 \\

        NeoVerse (2026) 
        & 50 & 157 & 207 & 0.855
        & 93.31
        & 92.70 & 94.74 & \textbf{60.19} & \textbf{65.26} & 96.84 & 99.01 
        & 0.069 & \textbf{0.998} \\
        \arrayrulecolor{gray}\midrule
        
        Inspatio-World (2026) 
        & 56.4 & 0.8 & \underline{57.2} & \underline{2.50}
        & \textbf{95.56}
        & \textbf{94.09} & \underline{95.16} & 59.92 & 61.54 & 97.56 & \underline{99.20} 
        & 0.055 & 1.731 \\
        
        \rowcolor{gray!10}
        \textbf{Ours} 
        & 0.9 & 0.98 & \textbf{1.88} & \textbf{10}
        & \underline{94.34}
        & 93.75 & 94.31 & 59.21 & \underline{62.19} & 97.54 & \textbf{99.33} 
        & \underline{0.053} & \underline{1.603} \\
        
        \arrayrulecolor{black}\bottomrule
    \end{tabular}%
    }
\end{table*}

%% file: tables/ablation_main_177.tex
\begin{table*}[t!]
\centering
\caption{\textbf{Ablation study in 177 frames.}
Rows (A--B) compare condition-injection strategies under the full pipeline; rows (C--F) ablate the two long-horizon designs on
top of Cross-Frame Injection.}
\vspace{-0.5em}
\label{tab:ablation-177}
\setlength\tabcolsep{3pt}
\resizebox{0.95\linewidth}{!}{%
\begin{tabular}{@{}l*{10}{c}@{}}
\toprule
\multirow{2}{*}[-1.33ex]{Exp.}
& \multirow{2}{*}[-1.33ex]{Condition Injection}
& \multirow{2}{*}[-1.33ex]{SR-RoPE}
& \multirow{2}{*}[-1.33ex]{CST}
& \multirow{2}{*}[-1.33ex]{CLIP-V $\uparrow$}
& \multicolumn{4}{c}{VBench~\citep{huang2024vbench} $\uparrow$}
& \multicolumn{2}{c}{Camera Acc. $\downarrow$} \\
\arrayrulecolor{gray}
\cmidrule(lr){6-9}
\cmidrule(lr){10-11}
& & & &
& \makecell{Sub.\\ Cons}
& \makecell{Bg.\\ Cons.}
& \makecell{Aes.\\ Qual.}
& \makecell{Img.\\ Qual.}
& \makecell{Trans\\ Err}
& \makecell{Rot\\ Err} \\
\midrule
(A) & Temporal Concat. & \checkmark & \checkmark
& 90.65
& 91.45 & 93.48 & 56.72 & 56.55 & 0.080 & 2.931 \\

(B) & Channel-wise Concat. & \checkmark &\checkmark
& 84.11
& 89.13 & 91.59 & 56.46 & 57.37 & 0.094 & 2.920 \\
\arrayrulecolor{gray}\midrule

(C) & Cross-Frame (Ours)   & $\times$   & $\times$   & 92.70
& 91.14 & 93.78 & 58.80 & 58.29 & 0.091 & 7.105 \\

(D) & Cross-Frame (Ours)  & \checkmark & $\times$   & 93.83
& 93.35 & 94.26 & 58.74 & 60.63 & 0.074 & 3.636 \\

(E) & Cross-Frame (Ours) & $\times$  & \checkmark
& 93.80
& 90.73 & 93.72 & 59.19 & 60.31 & 0.068 & 2.487 \\

\rowcolor{gray!10}
(F) & Cross-Frame (Ours) & \checkmark & \checkmark & {94.34}
& {93.75} & {94.31} & {59.21} & {62.19} & {0.053} & {1.603} \\
\arrayrulecolor{black}\bottomrule
\end{tabular}%
}
\end{table*}

%% file: sec/5_conclusion.tex
\section{Conclusion}
We presented \emph{LiveView}, a real-time framework for
camera-controlled video re-filming from continuously arriving video
streams. Our teacher-to-student framework combines Cross-Frame Injection for
synchronized source--target conditioning, Shared Relative RoPE for
bounded temporal indexing, and Cycle Streaming Tuning for stable
long-horizon autoregressive generation. This design enables the model to
preserve source content and dynamics while following online camera controls
without offline scene reconstruction. Experiments on short and 177-frame
videos demonstrate competitive visual quality and camera accuracy, while
achieving 10 FPS streaming rendering and substantially lower startup latency
than existing methods. These results establish LiveView as a
practical step toward interactive video-conditioned world models and
immersive real-time ``walk-through'' experiences. Limitations and future directions are discussed in the Appendix.